\documentclass[runningheads]{llncs}
\usepackage{graphicx}
\usepackage{amsmath,amssymb} 
\usepackage{color}
\usepackage{enumerate}
\usepackage[colorlinks]{hyperref}
\usepackage{booktabs}
\usepackage{helvet}
\usepackage{array}

\DeclareMathOperator*{\argmax}{arg\,max}

\begin{document}
%
\title{Video Re-localization 
} 

\titlerunning{Video Re-localization}
%
\author{Yang Feng$^\ddagger$\thanks{This work was done while Yang Feng was a Research Intern with Tencent AI Lab.} \quad
Lin Ma$^\dagger$ \quad
Wei Liu$^\dagger$ \quad
Tong Zhang$^\dagger$ \quad
Jiebo Luo$^\ddagger$
}
%
\authorrunning{Yang Feng, Lin Ma, Wei Liu, Tong Zhang, Jiebo Luo}
%

\institute{$^\dagger$Tencent AI Lab \quad $^\ddagger$University of Rochester\\
\email{\{yfeng23,jluo\}@cs.rochester.edu, forest.linma@gmail.com, wl2223@columbia.edu, tongzhang@tongzhang-ml.org}
}
\maketitle              
\begin{abstract}
Many methods have been developed to help people find the video contents they want efficiently. However, there are still some unsolved problems in this area. For example, given a query video and a reference video, how to accurately localize a segment in the reference video such that the segment semantically corresponds to the query video? We define a distinctively new task, namely \textbf{video re-localization}, to address this scenario. Video re-localization is an important emerging technology implicating many applications, such as fast seeking in videos, video copy detection, video surveillance, etc. Meanwhile, it is also a challenging research task because the visual appearance of a semantic concept in videos can have large variations. The first hurdle to clear for the video re-localization task is the lack of existing datasets. It is labor expensive to collect pairs of videos with semantic coherence or correspondence and label the corresponding segments. We first exploit and reorganize the videos in ActivityNet to form a new dataset for video re-localization research, which consists of about 10,000 videos of diverse visual appearances associated with localized boundary information. Subsequently, we propose an innovative cross gated bilinear matching model such that every time-step in the reference video is matched against the attentively weighted query video. Consequently, the prediction of the starting and ending time is formulated as a classification problem based on the matching results. Extensive experimental results show that the proposed method outperforms the competing methods. Our code is available at: \url{https://github.com/fengyang0317/video_reloc}.

\end{abstract}
\section{Introduction}
A great number of videos are generated every day. To effectively access the videos, several kinds of methods have been developed. The most common and mature one is searching by keywords. 
However, keyword-based search largely depends on user tagging. The tags of a video are user specified and it is unlikely for a user to tag all the content in a complex video. Content-based video retrieval (CBVR) \cite{chang1998fully,ren2009state,hu2011survey} has emerged to circumvent the tagging issue. Given a query video, CBVR systems analyze the content in it and retrieve videos with relevant contents to the query video. After retrieving videos, the user will have many videos in hand. It is time-consuming to watch all the videos from the beginning to the end to determine the relevance. Thus, video summarization methods \cite{zhang2016video,plummer2017enhancing} have been proposed to create a brief synopsis of a long video. Users are able to get the general idea of a long video quickly with the help of video summarization. Similar to video summarization, video captioning \cite{wang2018reconstruction,wang2018bidirectional} aims to summarize a video using one or more sentences. Researchers have also developed localization methods to help users quickly seek some video clips in a long video. The localization methods mainly focus on localizing video clips belonging to a list of pre-defined classes, for example, actions \cite{shou2017cdc,kalogeiton2017action}. Recently, localization methods with natural language queries have been developed \cite{Hendricks_2017_ICCV,Gao_2017_ICCV}.

\begin{figure}[t]
\centering
\includegraphics[width=\textwidth]{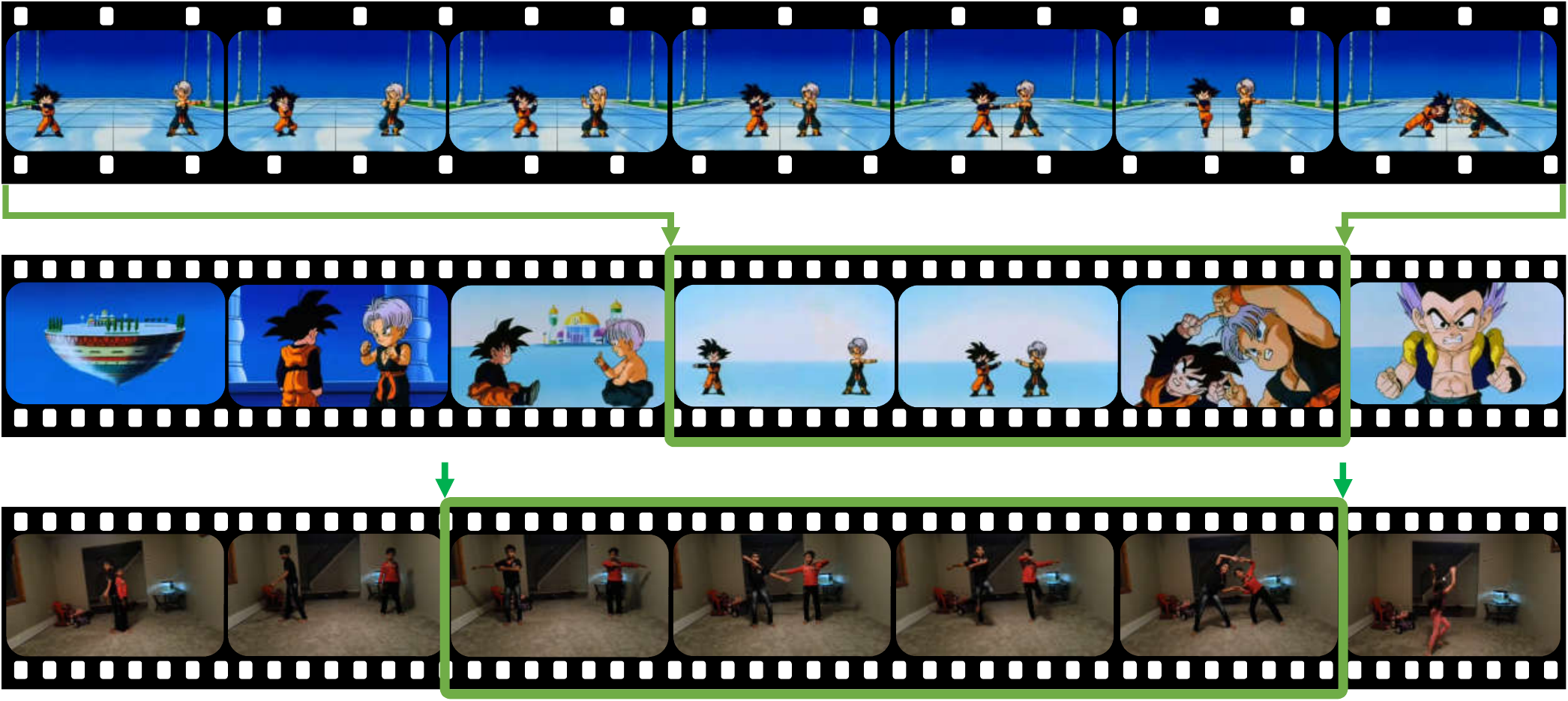}
\caption{The top video is a clip of an action performed by two characters. The middle video is a whole episode which contains the same action happening in a different environment (marked by a green rectangle). The bottom is a video containing the same action but performed by two real persons. Given the top query video, video re-localization aims to accurately detect the starting and ending points of the green segments in the middle and bottom video, respectively. Such segments semantically correspond to the given query video.}
\label{fig1}
\end{figure}

Although existing video retrieval techniques are powerful, there still remain some unsolved problems. Let us consider the following scenario: when a user is watching YouTube, he or she finds a very interesting video clip as shown in the top row of Fig. \ref{fig1}. This clip shows an action performed by two boy characters in a cartoon named ``\texttt{Dragon Ball Z}''. What should the user do if he/she wants to find when such an action also happens in that cartoon? Simply finding exactly the same content using copy detection methods \cite{jiang2016partial} would fail for most cases, as the content variations across videos are of great differences. As shown in the middle video of Fig.~\ref{fig1}, the action takes place in a different environment. Copy detection methods cannot handle such complicated scenarios. An alternative approach is relying on a proper action localization method. However, action localization methods usually localize pre-defined actions. When the action within the video clip, as shown in Fig. \ref{fig1}, has not been pre-defined or seen in the training dataset, action localization methods will not work. Therefore, an intuitive way to solve this problem is to crop the segment of interest as the query video and design a new model to localize the semantically matched segments in full episodes.

Motivated by this example, we define a distinctively new task called video re-localization, which aims at localizing a segment in a reference video such that this segment semantically corresponds to a query video. Specifically, the inputs to the new task are one query video and one reference video. The query video is a short clip which users are interested in. The reference video contains at least one segment semantically corresponding to the content in the query video. Then, video re-localization aims at accurately detecting the starting and ending points of the segment, which semantically corresponds to the query video.


Video re-localization implicates many real applications. With a query clip, a user can quickly find the content he/she is interested in by video re-localization, thus avoiding seeking in a long video manually. Video re-localization can also be applied to video surveillance and video-based person re-identification \cite{liu2017neural,liu2017video}. 

Video re-localization is a very challenging task. First, the appearances of the query and reference videos may be quite  different due to environment, subject, and viewpoint variances, even though they express the same visual concept. 
Second, determining the accurate starting and ending points is very challenging. There may be no obvious boundaries at the starting and ending points. Another key obstacle to video re-localization is the lack of video datasets that contain pairs of query and reference videos as well as the associated localization information. 

In order to tackle the video re-localization problem, we create a new dataset by reorganizing the videos in ActivityNet \cite{caba2015activitynet}. When building the dataset, we assume that the action segments belonging to the same class semantically correspond to each other. The query video is the segment that contains one action. The paired reference video contains one segment of the same type of action and the background information before and after the segment. We randomly split the 200 action classes into three parts. 160 action classes are used for training and 20 action classes are used for validation. The remaining 20 action classes are used for testing. Such a split guarantees that the action class of a video used for testing is unseen during training. Therefore, if the performance of a video re-localization model is good on the testing set, it should be able to generalize to other unseen actions as well.

To address the technical challenges of video re-localization, we propose a cross gated bilinear matching model of three recurrent layers. First, local video features are extracted from both the query and reference videos. The feature extraction is performed considering only a short period of video frames. The first recurrent layer is used to aggregate the extracted features and generate a new video feature considering the context information. Based on the aggregated representations, we perform matching between the query and reference videos. The feature of every reference video is matched with the attentively weighted query video. In  each matching step, the reference video feature and the query video feature are processed by factorized bilinear matching to generate their interaction results. Since not all the parts in the reference video are equally relevant to the query video, a cross gating strategy is stacked before bilinear matching to preserve the most relevant information while gating out the irrelevant information. The obtained interaction results are fed into the second recurrent layer to generate a query-aware reference video representation. The third recurrent layer is used to perform localization, where the prediction of the starting and ending positions is formulated as a classification problem. For each time step, the recurrent unit outputs the probability that the time step belongs to one of the four classes: starting point, ending point, inside the segment, and outside the segment. The final prediction result is the segment with the highest joint probability in the reference video. 

In summary, our contributions are four-fold:
\begin{enumerate}[1.]
	\item We introduce a novel task, namely video re-localization, which aims at localizing a segment in the reference video such that the segment semantically corresponds to the given query video.
    \item We reorganize the videos in ActivityNet \cite{caba2015activitynet} to form a new dataset to facilitate the research on video re-localization.
    \item We propose a cross gated bilinear matching model with the video re-localization task formulated as a classification problem, which can comprehensively capture the interactions between the query and reference videos. 
    \item We validate the effectiveness of our proposed model on the new dataset and achieve favorable performance better than the competing methods.
\end{enumerate}



\section{Related Work}

CBVR systems \cite{chang1998fully,ren2009state,hu2011survey} have evolved for two decades. 
Modern CBVR systems support various types of queries, such as query by example, query by object, query by keyword, and query by natural language. Given a query, CBVR systems can retrieve a list of entire videos related to the query. Some of the retrieved videos will inevitably contain contents irrelevant to the query. Users may still need to manually seek the part of interest in a retrieved video, which is time-consuming. Video re-localization introduced in this paper is different from CBVR in the sense that the former can locate the exact starting and ending points of the semantically coherent segment in a long reference video.


Action localization \cite{lan2011discriminative,klaser2010human} is related to video re-localization in the sense that both are intended to find the starting and ending points of a segment in a long video. The difference is that action localization methods merely focus on certain pre-defined action classes. Some attempts were made to go beyond pre-defined classes. Seo \textit{et al.} \cite{seo2011action} proposed a one-shot action recognition method that does not require prior knowledge about actions. 
Soomro and Shah \cite{soomro2017unsupervised} moved one step further by introducing unsupervised action discovery and localization. In contrast, video re-localization is more general than one-shot or unsupervised action localization in the sense that video re-localization can be applied to many other concepts besides actions.

Recently, Hendricks \textit{et al.} \cite{Hendricks_2017_ICCV}  proposed to retrieve a specific temporal segment from a video by a natural language query. 
Gao \textit{et al.} \cite{Gao_2017_ICCV} focused on temporal localization of actions in untrimmed videos using natural language queries. Compared to existing action localization methods, they have the advantage of localizing more complex actions than those in a pre-defined list. Our method is different in the sense that we directly match the query and reference video segments in a single video modality. 

\section{Methodology}
\label{sec:method}
\begin{figure}[t]
\centering
\includegraphics[width=\textwidth]{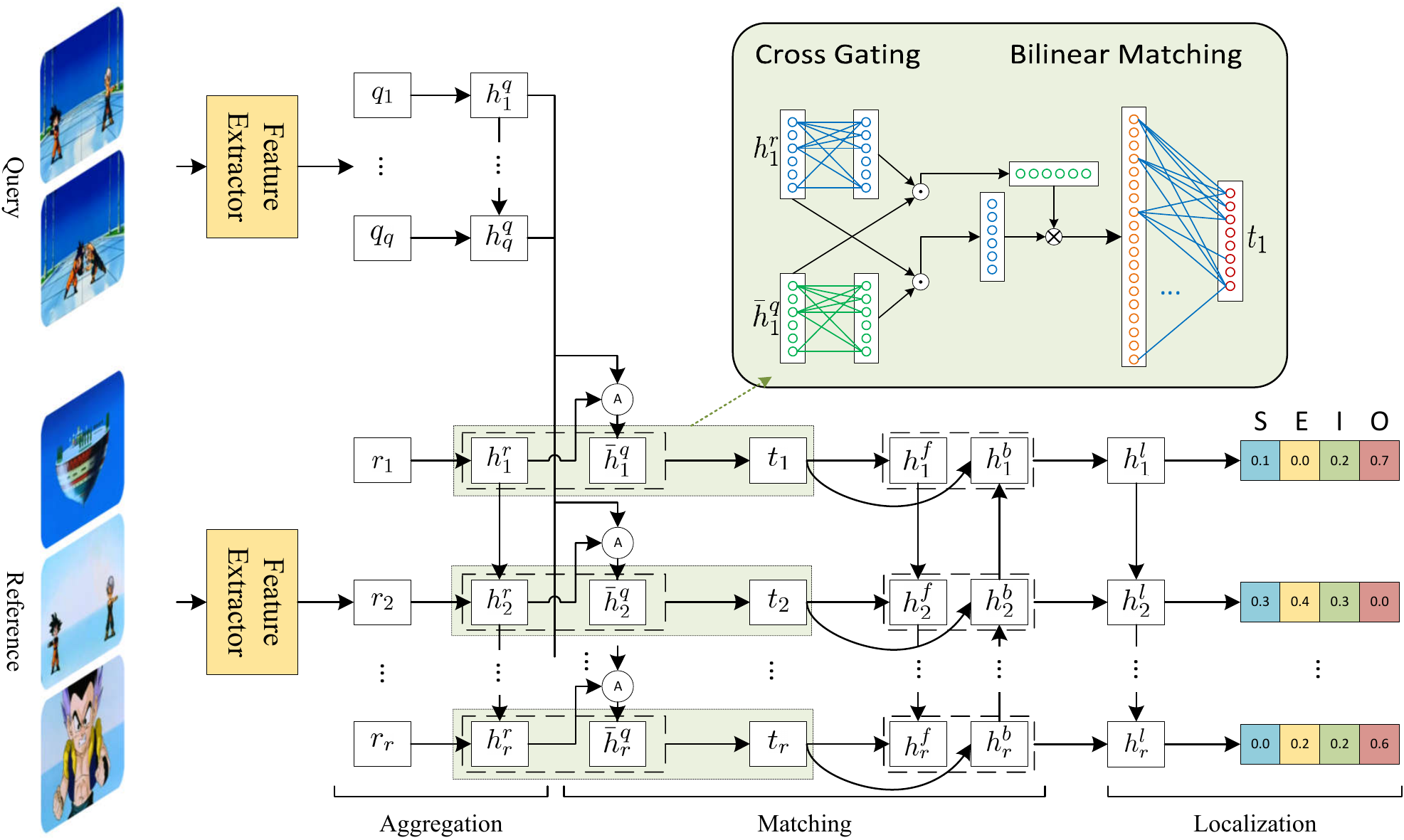}
\caption{The architecture of our proposed model for video re-localization. Local video features are first extracted for both query and reference videos and then aggregated by LSTMs. The proposed cross gated bilinear matching scheme exploits the complicated interactions between the aggregated query and reference video features. The localization layer, relying on the matching results, detects the starting and ending points of a segment in the reference video by performing classification on the hidden state of each time step. The four possible classes are \textbf{S}tarting, \textbf{E}nding, \textbf{I}nside and \textbf{O}utside. \textcircled{{\tiny\fontfamily{phv}\selectfont A}} denotes the attention mechanism described in Sec. \ref{sec:method}. $\odot$ and $\otimes$ are inner and outer products, respectively.}
\label{fig2}
\end{figure}

Given a query video clip and a reference video, we design a novel model to address the video re-localization task by exploiting their complicated interactions and predicting the starting and ending points of the matched segment. As shown in Fig.~\ref{fig2}, our model consists of three components, which are aggregation, matching, and localization. 

\subsection{Video Feature Aggregation}
In order to effectively represent video contents, we need to choose one or several kinds of video features depending on what kind of semantics we intend to capture. For our video re-localization task, the global video features are not considered, as we need to rely on the local information to perform segment localization. 

After performing feature extraction, two lists of local features with a temporal order are obtained for the query and reference videos, respectively. The query video features are denoted by a matrix $Q\in\mathbb{R}^{d\times q}$, where $d$ is the feature dimension and $q$ is the number of features in the query video, which is related to the video length. Similarly, the reference video is denoted by a matrix $R\in\mathbb{R}^{d\times r}$, where $r$ is the number of features in the reference video. As aforementioned, feature extraction only considers video characteristics within a short range. In order to incorporate contextual information within a longer range, we employ two long short-term memory (LSTM) \cite{hochreiter1997long} units to aggregate the extracted features: 
\begin{equation}
\begin{aligned}
h^q_i &= \text{LSTM}(q_i, h^q_{i-1}), \\
h^r_i &= \text{LSTM}(r_i, h^r_{i-1}),
\end{aligned}
\end{equation}
where $q_i$ and $r_i$ are the $i$-th columns in $Q$ and $R$, respectively. $h_i^q$, $h_i^r\in \mathbb{R}^{l\times 1}$ are the hidden states at the $i$-th time step of the two LSTMs, with $l$ denoting the dimensionality of the hidden state. Note that the parameters of the two LSTM are shared to reduce the model size. The yielded hidden states of the LSTMs are regarded as new video representations. Due to natural characteristics and behaviors of LSTMs, the hidden states can encode and aggregate the previous contextual information.

\subsection{Cross Gated Bilinear Matching}

At each time step,  we perform matching of the query and reference videos, based on the aggregated video representations $h^q_i$ and $h^r_i$. Our proposed cross gated bilinear matching scheme consists of four modules, i.e., the generation of attention weighted query, cross gating, bilinear matching, and matching aggregation.

\subsubsection{Attention Weighted Query.} For video re-localization, the segment corresponding to the query clip can potentially be anywhere in the reference video. Therefore, every feature from the reference video needs to be matched against the query video to capture their semantic correspondence. Meanwhile, the query video may be quite long, so only some parts in the query video actually correspond to one feature in the reference video. Motivated by the machine comprehension method in \cite{wang2016machine}, an attention mechanism is leveraged to select which part in the query video is to be matched with the feature in the reference video. At the $i$-th time step of the reference video, the query video is weighted by the attention mechanism:
\begin{equation}
\label{eq:att}
\begin{aligned}
e_{i,j} &= \tanh(W^qh^q_j + W^rh^r_i + W^mh^f_{i-1} + b^m), \\
\alpha_{i,j} &= \frac{\exp(w^\top e_{i,j} + b)}{\sum_k \exp(w^\top e_{i,k} + b)}, \\
\bar{h}^q_i &= \sum_j \alpha_{i,j}h^q_j,
\end{aligned}
\end{equation}
where $W^q$, $W^r$, $W^m\in\mathbb{R}^{l\times l}$, $w\in\mathbb{R}^{l\times 1}$ are the weight parameters in our attention model with $b^m\in\mathbb{R}^{l\times 1}$ and $b\in\mathbb{R}$ denoting the bias terms. It can be observed that the attention weight $\alpha_{i,j}$ relies on not only the current representation $h_i^r$ of the reference video but also the matching result $h_{i-1}^f\in\mathbb{R}^{l\times 1}$ in the previous stage, which can be obtained by Eq. (\ref{eq:match_lstm}) and will be introduced later. The attention mechanism tries to find the most relevant $h_j^q$ to $h_i^r$ and use the relevant $h_j^q$ to generate the query representation $\bar{h}_i^q$, which is believed to better match $h_i^r$ for the video re-localization task. 

\subsubsection{Cross Gating.}
 Based on the attention weighted query representation $\bar{h}_i^q$ and reference representation $h_i^r$, we propose a cross gating mechanism to gate out the irrelevant reference parts and emphasize the relevant parts. In cross gating, the gate for the reference video feature depends on the query video. Meanwhile, the query video features are also gated by the current reference video feature. The cross gating mechanism can be expressed by the following equation:
\begin{equation}
\begin{aligned}
g_i^r = \sigma(W^g_r h^r_i + b^g_r)&,\qquad \tilde{h}_i^q = \bar{h}_i^q\odot g_i^r, \\
g_i^q = \sigma(W^g_q \bar{h}^q_i + b^g_q)&,\qquad \tilde{h}_i^r = h^r_i\odot g_i^q,
\end{aligned}
\end{equation}
where $W^g_r, W^g_q\in\mathbb{R}^{l\times l}$, and $b^g_r, b_q^g\in\mathbb{R}^{l\times 1}$ denote the learnable parameters. $\sigma$ denotes the non-linear sigmoid function. If the reference feature $h_i^r$ is irrelevant to the query video, both the reference feature $h_i^r$ and query representation $\bar{h}_i^q$ are filtered to reduce their effects on the subsequent layers. If $h_i^r$ closely relates to $\bar{h}_i^q$, the cross gating strategy is expected to further enhance their interactions.

\subsubsection{Bilinear Matching.} Motivated by bilinear CNN \cite{lin2015bilinear}, we propose a bilinear matching method to further exploit the interactions between $\tilde{h}^q_i$ and $\tilde{h}^r_i$, which can be written as:
\begin{equation}
\label{eq:bilinear_old}
t_{ij} = \tilde{h}^{q\top}_iW^b_j\tilde{h}^r_i + b_j^b,
\end{equation}
where $t_{ij}$ is the $j$-th dimension of the bilinear matching result, given by $t_i=[t_{i1},t_{i2},\ldots,t_{il}]^\top$. $W^b_j\in\mathbb{R}^{l\times l}$ and $b^b_j\in\mathbb{R}$ are the learnable parameters used to calculate $t_{ij}$.

The bilinear matching model in Eq. (\ref{eq:bilinear_old}) introduces too many parameters, thus making the model difficult to learn. Normally, to generate an $l$-dimension bilinear output, the number of parameters introduced would be $l^3 + l$. In order to reduce the number of parameters, we factorize the bilinear matching model as:
\begin{equation}
\label{eq:bilinear}
\begin{aligned}
\hat{h}^q_i &= F_j\tilde{h}^q_i + b^f_j, \\
\hat{h}^r_i &= F_j\tilde{h}^r_i + b^f_j, \\
t_{ij} &= \hat{h}^{q\top}_i\hat{h}^r_i, \\
\end{aligned}
\end{equation}
where $F_j\in\mathbb{R}^{k\times l}$ and $b^f_j\in\mathbb{R}^{k\times 1}$ are the parameters to be learned. $k$ is a hyper-parameter much smaller than $l$. Therefore, only $k\times l\times(l+1)$ parameters are introduced by the factorized bilinear matching model.

The factorized bilinear matching scheme captures the relationships between the query and reference representations. By expanding Eq. (\ref{eq:bilinear}), we have  the following equation:
\begin{equation}
t_{ij} = \underbrace{\tilde{h}^{q\top}_iF_j^\top F_j\tilde{h}^r_i}_{\text{quadratic\ term}} +
\underbrace{ b^{f\top}_jF_j(\tilde{h}^q_i + \tilde{h}^r_i)}_{\text{linear term}} +
\underbrace{ b^{f\top}_ib^f_i}_{\text{bias term}}.
\end{equation}
Each $t_{ij}$ consists of a quadratic term, a linear term, and a bias term, with the quadratic term capable of capturing  the complex dynamics between $\tilde{h}_i^q$ and $\tilde{h}_i^r$.


\subsubsection{Matching Aggregation.} Our obtained matching result $t_i$ captures the complicated interactions between the query and reference videos from a local view point. Therefore, an LSTM unit is used to further aggregate the matching context:
\begin{equation}
\label{eq:match_lstm}
h^f_i = \text{LSTM}(t_i, h^f_{i-1}).
\end{equation}
Following the idea in bidirectional RNN \cite{schuster1997bidirectional}, we also use another LSTM unit to aggregate the matching results in the reverse direction. Let $h^b_i$ denote the hidden state of the LSTM in the reverse direction. By concatenating $h^f_i$ together with $h^b_i$, the aggregated hidden state $h^m_i$ is generated.

\subsection{Localization}
The output of the matching layer $h^m_i$ indicates whether the content in the $i$-th time step in the reference video matches well with the query clip. We rely on $h_i^m$ to predict the starting and ending points of the matching segment. Specifically, we formulate the localization task as a classification problem. As illustrated in Fig.~\ref{fig2}, at each time step in the reference video, the localization layer predicts the probability that this time step belongs to one of the four classes: starting point, ending point, inside point, and outside point. The localization layer is given by:
\begin{equation}
\begin{aligned}
h^l_i &= \text{LSTM}(h^m_i, h^l_{i-1}), \\
p_i &= \text{softmax}(W^l h^l_i + b^l),
\end{aligned}
\end{equation}
where $W^l\in\mathbb{R}^{4\times l}$ and $b^l\in\mathbb{R}^{4\times 1}$ are the parameters in the softmax layer. $p_i$ is the predicted probability for time step $i$. It has four dimensions $p_{i}^1$, $p_{i}^2$, $p_{i}^3$, and $p_{i}^4$, denoting the probability of starting, ending, inside and outside, respectively.

\subsection{Training}
We train our model using the weighted cross-entropy loss. We generate a label vector for the reference video at each time step. For a reference video with a ground-truth segment $[s,e]$, we assume $1\leq s\leq e\leq r$. The time steps belonging to $[1,s)$ and $(e,r]$ are outside the ground-truth segment, and the generated label probabilities for them are $g_i=[0,0,0,1]$. The $s$-th time step is the starting time step, which is assigned to label probabilities $g_i=[\frac{1}{2},0,\frac{1}{2},0]$. Similarly, the label probabilities at the $e$-th time step are $g_i=[0,\frac{1}{2},\frac{1}{2},0]$. The time steps in the segment $(s,e)$ are labeled as $g_i=[0,0,1,0]$. When the segment is very short and falls in only one time step, $s$ will be equal to $e$. In that case, the label probabilities for that time step would be $[\frac{1}{3}, \frac{1}{3}, \frac{1}{3}, 0]$. The cross-entropy loss for one sample pair is given by:
\begin{equation}
loss = -\frac{1}{r}\sum_{i=1}^r\sum_{n=1}^4g_{i}^n\log(p_{i}^n),
\end{equation}
where $g_i^n$ is the $n$-th dimension of $g_i$.

One issue of using the above loss for training is that the predicted probabilities of the starting point and ending point would be orders smaller than the probabilities of the other two classes. The reason is that the positive samples for the starting and ending points are much fewer than those of the other two classes. For one reference video, there is only one starting point and one ending point. In contrast, all the other positions are either inside or outside the segment. Hence, we decide to pay more attention to losses at the starting and ending positions, via a dynamic weighting strategy:
\begin{equation}
w_i = \left\{
\begin{array}{ll}
c_w,\ \ \ & \textrm{if $g_{i}^1 + g_{i}^2 > 0$}, \\
\ 1, & \textrm{otherwise},
\end{array} \right.
\end{equation}
where $c_w$ is a constant. Thus, the weighted loss used for training can be further formulated as:
\begin{equation}
loss^w = -\frac{1}{r}\sum_{i=1}^r w_i\sum_{n=1}^4g_{i}^n\log(p_{i}^n).
\end{equation}

\subsection{Inference}
After the model is properly trained, we can perform video re-localization on a pair of query and reference videos. We localize the segment with the largest joint probability in the reference video, which is given by:
\begin{equation}
\label{eq:pred}
s,e = \argmax_{s,e}p_{s}^1p_{e}^2\left(\prod_{i=s}^ep_{i}^3\right)^{\frac{1}{e-s+1}},
\end{equation}
where $s$ and $e$ are the predicted time steps of the starting and ending points, respectively. As shown in Eq.~(\ref{eq:pred}), the geometric mean of all the probabilities inside the segment is used such that the joint probability will not be affected by the length of the segment. 

\section{The Video Re-localization Dataset}
Existing video datasets are usually created for classification \cite{kay2017kinetics,THUMOS15}, temporal localization \cite{caba2015activitynet}, captioning \cite{chen2011collecting}, or video summarization  \cite{gygli2014creating}. None of them can be directly used for the video re-localization task. 
To train our video re-localization model, we need pairs of query videos and reference videos, where the segment in the reference video semantically corresponding to the query video should be annotated with its localization information, specifically the starting and ending points. 
It would be labor expensive to manually collect query and reference videos and localize the segments sharing the same semantics as the query video.

\begin{figure}[t]
\centering
\includegraphics[width=\textwidth]{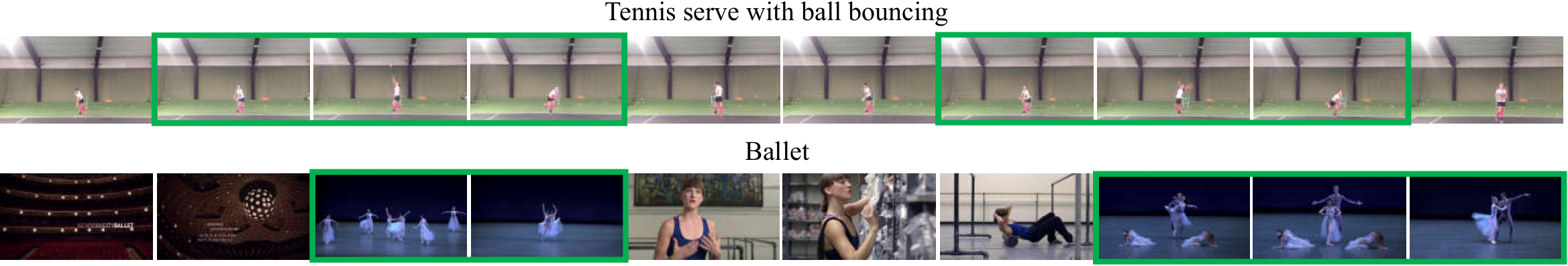}
\caption{Several video samples in our dataset. The segments containing different actions are marked by green rectangles.}
\label{fig:sample}
\end{figure}


Therefore, in this work, we create a new dataset based on ActivityNet \cite{caba2015activitynet} for video re-localization. ActivityNet is a large-scale action localization dataset with segment-level action annotations. We reorganize the video sequences in ActivityNet aiming to relocalize the actions in one video sequence given another video segment of the same action. There are 200 classes in ActivityNet and the videos of each class are split into training, validation and testing subsets. This split is not suitable for our video re-localization problem, because we expect a video re-localization method that should be able to relocalize more actions than those actions defined in ActivityNet. Therefore, we split the dataset by action classes. Specifically, we randomly select 160 classes for training, 20 classes for validation, and the remaining 20 classes for testing. This split guarantees that the action classes used for validation and testing will not be seen during training. The video re-localization model is thus required to relocalize unknown actions during testing. If it works well on the testing set, it should be able to generalize well to other unseen actions.

Many videos in ActivityNet are untrimmed and contain multiple action segments. First, we filter the videos with two overlapped segments, which are annotated with different action classes. Second, we merge the overlapped segments of the same action class. Third, we also remove the segments that are longer than 512 frames. After such preprocessings, we obtain $9,530$ video segments. Fig.~\ref{fig:sample} illustrates several video samples in the dataset. It can be observed that some video sequences contain more than one segment. 
One video segment can be regarded as a query video clip, while its paired reference video can be selected or cropped from the video sequence to contain only one segment with the same action label as the query video clip. 
During our training process, the query video and reference video are randomly paired, while the pairs are fixed for validation and testing. In the future, we will release the constructed dataset to the public and continuously enhance the dataset.

\section{Experiments}
In this section, we conduct several experiments to verify our proposed model. First, three baseline methods are designed and introduced. Then we will introduce our experimental settings including evaluation criteria and implementation details. Finally, we demonstrate the effectiveness of our proposed model through performance comparisons and ablation studies.

\subsection{Baseline Models}
Currently, there is no model specifically designed for video re-localization. We design three baseline models, performing frame-level and video-level comparisons, and action proposal generation, respectively.

\subsubsection{Frame-level Baseline.} We design a frame-level baseline motivated by the backtracking table and diagonal blocks described in~\cite{chou2015pattern}. We first normalize the features of query and reference videos. Then we calculate a distance table $D\in\mathbb{R}^{q\times r}$ by $D_{ij} = \|h^q_i - h^r_j\|_2$. The diagonal block with the smallest average distances is searched by dynamic programming. The output of this method is the segment in which the diagonal block lies. Similar to~\cite{chou2015pattern}, we also allow horizontal and vertical movements to allow the length of the output segment to be flexible. Please note that no training is needed for this baseline.

\subsubsection{Video-level Baseline.} In this baseline, each video segment is encoded as a vector by an LSTM. The L2-normalized last hidden state in the LSTM is selected as the video representation. To train this model, we use the triplet loss in ~\cite{schroff2015facenet}, which enforces anchor positive distance to be smaller than anchor negative distance by a margin. The query video is regarded as the anchor. Positive samples are generated by sampling a segment in the reference video having temporal overlap (tIoU) over $0.8$ with the ground-truth segment while negative samples are obtained by sampling a segment with tIoU less than $0.2$. When testing, we perform exhaustively search to select the most similar segment with  the query video.

\subsubsection{Action Proposal Baseline.} We train the SST \cite{buch2017sst} model on our training set and perform the evaluation on the testing set. The output of the model is the proposal with the largest confidence score.

\subsection{Experimental Settings}
We use C3D \cite{tran2015learning} features released by ActivityNet Challenge 2016\footnote{\url{http://activity-net.org/challenges/2016/download.html}}. The features are extracted by publicly available pre-trained C3D model having a temporal resolution of 16 frames. The values in the second fully-connected layer (fc7) are projected to 500 dimensions by PCA. We temporally downsample the provided features by a factor of two so they do not have overlap with each other. Adam \cite{kingma2014adam} is used as the optimization method. The parameters for the Adam optimization method are left at defaults: $\beta_1=0.9$ and $\beta_2=0.999$. The learning rate, dimension of the hidden state $l$, loss weight $c_w$ and factorized matrix rank $k$ are set to 0.001, 128, 10, and 8, respectively. We manually limit the maximum allowed length of the predicted segment to 1024 frames.

Following the action localization task, we report the average top-1 mAP computed with tIoU thresholds between 0.5 and 0.9 with the step size of 0.1.



\begin{table}
\caption{Performance comparisons on our constructed dataset. The top entry in each column is highlighted in boldface.}
\label{table:res}
\centering
\begin{tabular}{l|>{\centering}m{1.2cm} >{\centering}m{1.2cm} >{\centering}m{1.2cm} >{\centering}m{1.2cm} >{\centering}m{1.2cm}|c}
\toprule
     mAP {\fontfamily{phv}\selectfont @}1 & 0.5 & 0.6  & 0.7  & 0.8  & 0.9  & Average \\ 
\hline
     Chance & 16.2 & 11.0 & 5.4 & 2.9 & 1.2 & 7.3 \\ 
     Frame-level baseline & 18.8 & 13.9 & 9.6 & 5.0 & 2.3 & 9.9 \\ 
     Video-level baseline & 24.3 & 17.4 & 12.0 & 5.9 & 2.2 & 12.4 \\ 
     SST \cite{buch2017sst} & 33.2 & 24.7 & 17.2 & 7.8 & 2.7 & 17.1 \\ 
     Our model & \textbf{43.5} & \textbf{35.1} & \textbf{27.3} & \textbf{16.2} & \textbf{6.5} & \textbf{25.7} \\ 
\bottomrule
\end{tabular}
\end{table}

\begin{figure}[t]
\centering
\includegraphics[width=9cm]{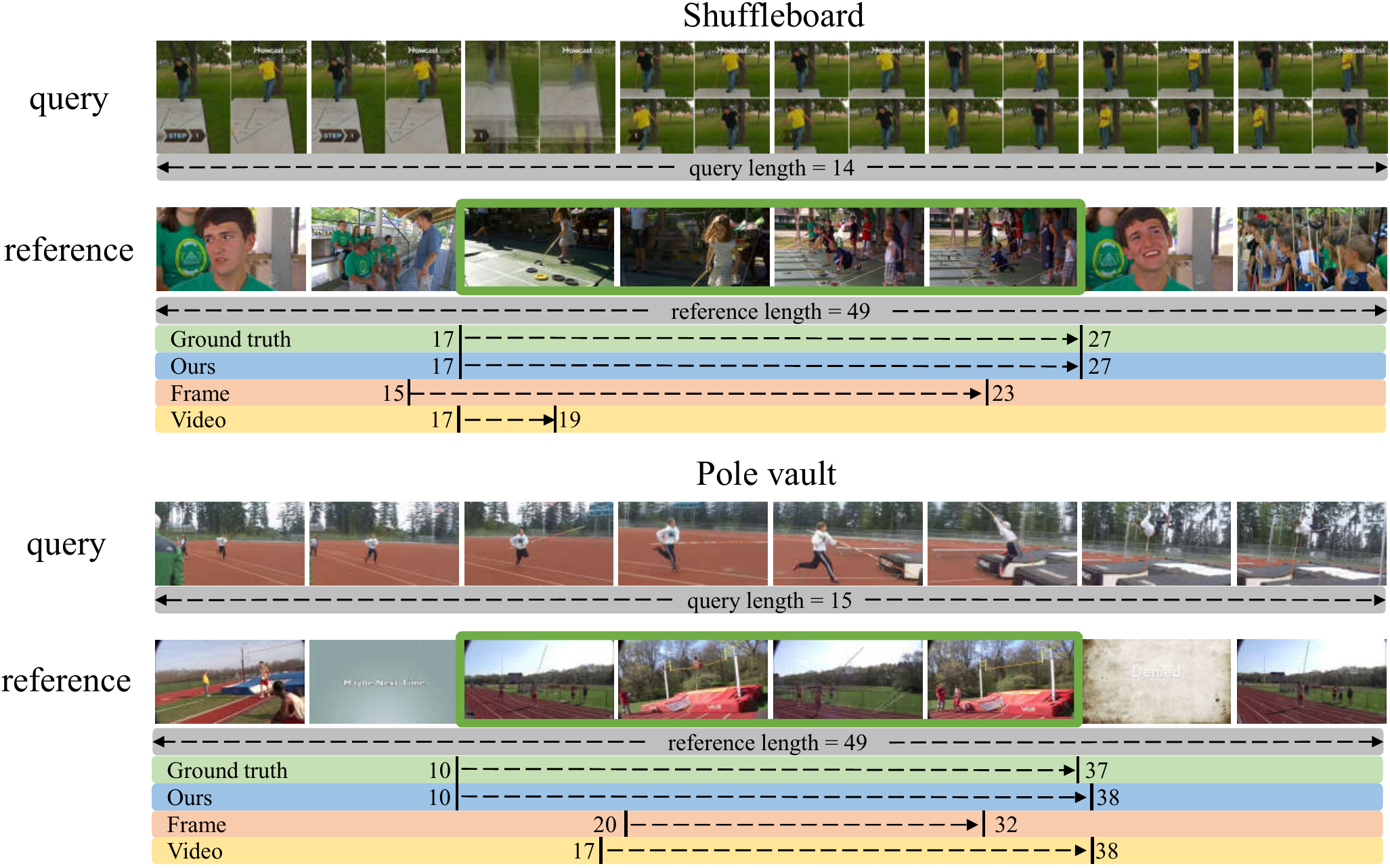}
\caption{Qualitative results. 
The segment corresponding to the query is marked by a green rectangle. Our model can accurately localize the segment  semantically corresponding to the query video in the reference video.}
\label{fig:qua}
\end{figure}

\begin{figure}[t]
\centering
\includegraphics[width=9cm]{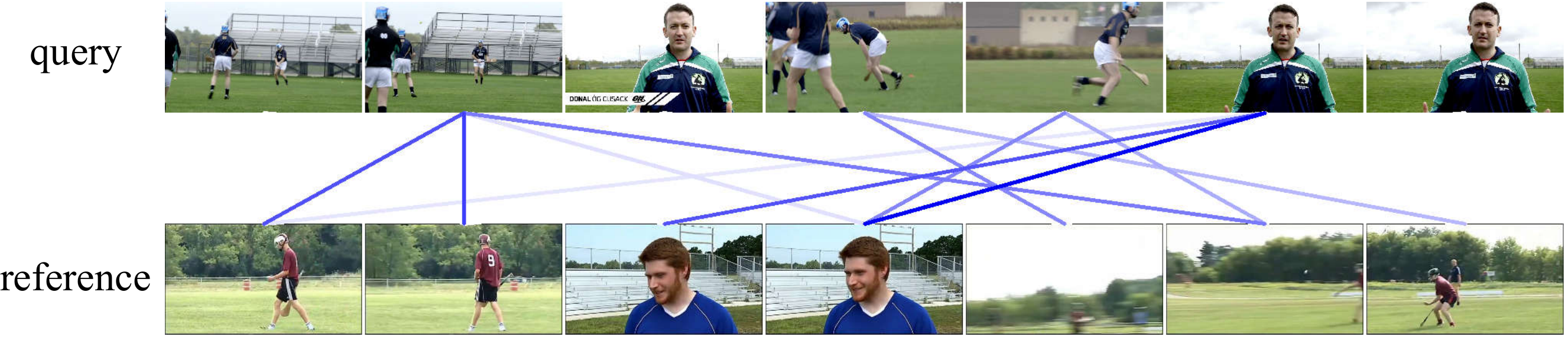}
\caption{Visualization of the attention mechanism. The top video is the query, while the bottom video is the reference. Color intensities of blue lines indicate the attention strengths. The darker the colors are, the higher the attention weights are. Note that only the connections with high attention weights are shown.}
\label{fig:att}
\end{figure}

\subsection{Performance Comparisons}
Table \ref{table:res} shows the results of our method and baseline methods. According to the results, we have several observations. The frame-level baseline performs better than randomly guesses, which suggests that the C3D features preserve the similarity between videos. The result of the frame-level baseline is significantly inferior to our model. The reasons may be attributed to the fact that no training process is involved in the frame-level baseline. 

The performance of the video-level baseline is slightly better than the frame-level baseline, which suggests that the LSTM used in the video-level baseline learns to project corresponding videos to similar representations. However, the LSTM encodes the two video segments independently without considering their complicated interactions. Therefore, it cannot accurately predict the starting and ending points. Additionally, this video-level baseline is very inefficient during the inference process because the reference video needs to be encoded multiple times for an exhaustive search.

Our method is substantially better than the three baseline methods. The good results of our method indicate that the cross gated bilinear matching scheme indeed helps to capture the interactions between the query and the reference videos. The starting and ending points can be accurately detected, demonstrating its effectiveness for the video re-localization task.

Some qualitative results from the testing set are shown in Fig. \ref{fig:qua}. It can be observed that the query and reference videos are of great visual difference, even though they express the same semantic meaning. Although our model has not seen these actions during the training process, it can effectively measure their semantic similarities, and consequently  localizes the segments correctly in the reference videos.

\begin{table}
\caption{Performance comparisons of the ablation study. The top entry in each column is highlighted in boldface.}
\label{table:abl}
\centering
\begin{tabular}{l|>{\centering}m{1.2cm} >{\centering}m{1.2cm} >{\centering}m{1.2cm} >{\centering}m{1.2cm} >{\centering}m{1.2cm}|c}
\toprule
	  mAP {\fontfamily{phv}\selectfont @}1 & 0.5 & 0.6 & 0.7 & 0.8 & 0.9 & Average \\ \hline
     Base & 40.8 & 32.4 & 22.8 & 15.9 & 6.4 & 23.7 \\ 
     Base + cross gating & 40.5 & 33.5 & 25.1 & 16.2 & 6.1 & 24.3 \\ 
     Base + bilinear & 42.3 & 34.9 & 25.7 & 15.4 & \textbf{6.5} & 25.0 \\ 
     Ours & \textbf{43.5} & \textbf{35.1} & \textbf{27.3} & \textbf{16.2} & \textbf{6.5} & \textbf{25.7} \\ 
\bottomrule
\end{tabular}
\end{table}

\subsection{Ablation Study}
\subsubsection{Contributions of Different Components.} To verify the contribution of each part of our proposed cross gated bilinear matching model, we perform three ablation studies. In the first ablation study, we create a base model by removing the cross gating part and replacing the bilinear part with the concatenation of two feature vectors. The second and third studies are designed by adding cross gating and bilinear to the base model, respectively. Table \ref{table:abl} lists all the results of the aforementioned ablation studies.
It can be observed that both bilinear matching and cross gating are helpful for the video re-localization task. Cross gating can help filter out the irrelevant information while  enhancing the meaningful interactions between the query and reference videos. Bilinear matching fully exploits the interactions between the reference and query videos, leading to better results than the base model. Our full model, consisting of both cross gating and bilinear matching, achieves the best results.



\subsubsection{Attention.} In Fig.~\ref{fig:att}, we visualize the attention values for a query and reference video pair. The top video is the query video, while the bottom video is the reference. Both of the two videos contain some parts of ``hurling" and ``talking". It is clear that the ``hurling'' parts in the reference video highly interact with the ``hurling" parts in the query with larger attention weights.

\section{Conclusions}
In this paper, we first defined a distinctively new task called video re-localization, which aims at localizing a segment in the reference video such that this segment semantically corresponds to the query video. Video re-localization implicates many real-world applications, such as finding interesting moments in videos, video surveillance, and person re-identification. To facilitate the new video re-localization task, we created a new dataset by reorganizing the videos in ActivityNet \cite{caba2015activitynet}. Furthermore, we proposed a novel cross gated bilinear matching method, which effectively performs the matching between the query and reference videos. Based on the matching results, an LSTM was applied to localize the query video in the reference video. Extensive experimental results show that our proposed model is effective and outperforms the baseline methods. 

\section*{Acknowledgement}
We would like to thank the support of New York State through the Goergen Institute for Data Science and NSF Award \#1722847.

%
%
%

\end{document}